# Semi-bounded Rationality: A model for decision making

Tshilidzi Marwala
P.O. Box 787391
Sandton, 2146
Republic of South Africa
E-mail: tmarwala@gmail.com

In this paper the theory of semi-bounded rationality is proposed as an extension of the theory of bounded rationality. In particular, it is proposed that a decision making process involves two components and these are the correlation machine, which estimates missing values, and the causal machine, which relates the cause to the effect. Rational decision making involves using information which is almost always imperfect and incomplete as well as some intelligent machine which if it is a human being is inconsistent to make decisions. In the theory of bounded rationality this decision is made irrespective of the fact that the information to be used is incomplete and imperfect and the human brain is inconsistent and thus this decision that is to be made is taken within the bounds of these limitations. In the theory of semi-bounded rationality, signal processing is used to filter noise and outliers in the information and the correlation machine is applied to complete the missing information and artificial intelligence is used to make more consistent decisions.

## 1 Introduction

Making decisions has been a complex task since time immemorial. In primitive society making decisions was muddled with superstitions (Vyse, 2000; Foster and Kokko, 2009). There has been even speculation that superstition also occurs in other animals such as pigeons (Skinner, 1948). In essence, superstition can be defined as supernatural causality where something is caused by another without them linked to one another. The idea of one event causing another without any connection between them whatsoever is characterized as irrational thinking.

   This paper proposes that the three fundamentals of decision making are: data analysis, which involves statistical analysis, correlation machine which is used to estimate missing data and causality, which relates inputs to outputs. As societies have made attempts to move away from superstition, the concept of rational decision making came to the fore. The concept of rationality has attracted many philosophers from different fields. Max Weber studied the role of rationality in social action (Weber, 1922) whereas Grayling studied the role of other social forces such as emotions on the quality or degree of rationality. Mosterin (2008) defined reason as a psychological facility and rationality as the optimizing strategy. In artificial intelligence a rational agent is that which aims to maximize its utility in making its decision. Utility here is defined as the usefulness of its decisions. Some researchers have extended the definition of rationality not only to include the use of logic and reason to make this decision but to say such decision should be optimized.

   Basically rational decision making is a process of reaching decisions through logic and reason (Nozick, 1993; Spohn, 2002). Decision making can be defined as a process through which a decision is reached. It usually involves many possible decisions outcomes and is said to be rational and optimal if it maximizes the good or utility that is derived from its consequences and at the same time minimizes the bad or the uselessness that is also derived from its consequences. The philosophical concept that states that the best course of action is that which maximizes utility is known as utilitarianism and was advocated by philosophers such as Jeremy Bentham and John Stuart Mills (Adams, 1976; Anscombe, 1958; Bentham, 2009; Mill, 2011).

There is a theory that has been proposed to achieve this desired outcome of a decision making process and this is the theory of rational choice and it states that one chooses a decision based on the product of the impact of the decision and its probability of occurrence (Allingham, 2002; Bicchieri, 2003). However, this theory was found to be inadequate because it does not take into account of where the person will be relative to where they initially were before making a decision. Kahneman (2011) extended the theory of rational choice by introducing the Prospect Theory which includes the reference position on evaluating the optimal decision (Kahneman and Tversky, 1979).

In this paper we propose a generalized decision making process with two sets of input information which form the basis for decision making and options that are available for making such decisions (model) and the output being the decision being made. It further proposes that within this generalized decision making framework lie the statistical analysis device, correlation machine and a causal machine. Thus the model proposed in this paper for decision making can be summarized as follows:

1. Optimization Device: e.g. maximizing utility and minimizing worthlessness which is what prospect theory does.
2. Decision Making Device: In this paper we propose that this element contains data analysis device, correlation machine and causal machine.
3. Control Device: In this paper it is proposed to be an open loop control system which entails merely calculating the utility (together with the uselessness) and choosing a decision which maximizes utility and minimize the uselessness.

The next section will describe in detail the concept of rational decision making, and the following section will describe the concept of bounded rationality which will be followed by the concept of semi-bounded rationality.

## 2 Rational Decision Making: A causal approach

Rational decision making in this paper is defined as a process through which a decision is made using some gathered information and some intelligence to make a decision. It is deemed to be rational because it is based on evidence in the form of information. Making decisions in the absence of any information is deemed irrational. Rational decision making as illustrated in Figure 1 entails using information and a model (e.g. human brain) to make a decision. The relationship between information or data and the impact of the decision is used to improve or reinforce the learning process.

Baker (2013) conducted a case study on rational decision making in a hybrid organization for a Southwestern Drug Court treatment program. Serrano González *et. al.* (2013) successfully applied rational decision making to identify optimal structure of transmissions systems for offshore wind farms. Salem *et. al.* (2013) applied successfully rational multi-criteria decision making to choose an operational plan for bridge rehabilitation. A multi-criteria optimization is an optimization of problem with more than one objective (Steuer, 1986; Köksalan *et. al.,* 1995; Zionts and Wallenius, 1976). As an example, one is given an option to buy a certain number of Microsoft stocks and General Electric stocks and the decision is how one combines these two stocks.

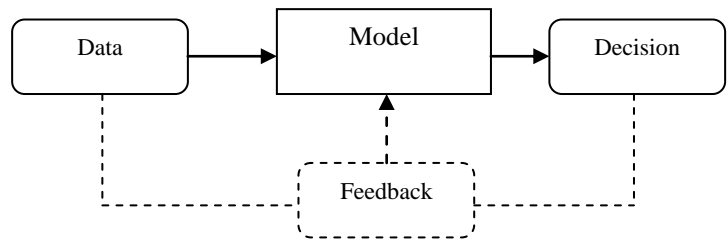

Figure 1 An illustration of a rational decision making system

Chai *et. al.* (2013) reviewed how rational decision making has been applied successfully to choose a supplier selection whereas Badri *et. al.* (2013) applied rational decision making successfully to identify fireproofing requirements against jet fires. Other studies on the subject include studying a relationship between subjective memory and rational decision making (Hembacher and Ghetti, 2013) as well as successful application of rational decision making to economics (Gu *et. al.,* 2013).

## 3 Bounded-Rational Decision Making

One definition of rationality includes the concept of making an optimized decision. Practically this is not possible because of the limitation of the availability of information required to make such a decision and the limitation of the device to make sense of such incomplete decision (Simon, 1990). This is what is known as the theory of bounded rationality and was proposed by the Nobel Laureate Herbert Simon (Simon, 1957; Tisdell, 1996). The implication of this observation on many economic systems is quite substantial (Simon, 1991). It questions the truthfulness of the theory of the efficient market hypothesis especially given the fact that participants in the market are at best only rational within the bounds of limited information and limited model to make sense of the information.

Lee (2013) built a model for analyzing an economy using bounded rational agents. It was observed that rational agents amplify of the price bubble cooperation amongst rational agents. Gama (2013) successfully applied the concept of bounded rationality to investigate the behavior of stream mining procedures and proposed ubiquitous stream mining and self-adaption models. Jiang *et. al.* (2013) successfully applied bounded rationality in evolution game analysis of water saving and pollution prevention. Jin *et. al.* (2013) successfully applied the theory of bounded rationality and game theory to construct a computer virus propagation model and observed that the proposed model was able to forecast the propagation of computer virus.

Yao and Li (2013) surmised that the concept of bounded rationality can be viewed as a basis of loss aversion and optimism and in this regard studied psychological adaptation with the context of incomplete information. They observed that loss aversion and optimism arise when the degree of information incompleteness exceeds a particular threshold. In addition, they observed that loss aversion and optimism develop to be more noticeable when information is sparser. They then concluded that psychological biases benefit from apparent information incompleteness when value creation is considered.

Stanciu-Viziteu (2012) simulated the behavior of sharks to model financial market where investors are embodied by hungry sharks. The results obtained indicated that sharks resembling investors coordinate and produce equilibrium under rational expectations.

Aviad and Roy (2012) applied the concept of bounded rationality to build a decision support technique and applied this to identify feature saliency in clustering problems. Bounded rationality was applied to deal with the problem of the availability of target attribute by using an S-shaped function as a saliency measure to characterize the end user's logic to identify attributes that describe each prospective group.

Murata *et. al.* (2012) applied the concept of bounded rationality to analyze the relationship between group heuristics and cooperative behavior. They observed that group consciousness can assist to stimulate actively mutual cooperation.

The literature review studied above clearly indicates that the theory of bounded rationality is a powerful tool of analysis that has found usage in many diverse areas. It should be noted that the theory of bounded rationality has not replaced the theory of rationality which was described in earlier section. What this theory has done is to put limitations to the applicability of the theory of rationality. The theory of rationality which was described in Figure 1 can thus be updated to construct the theory of bounded rationality as described in Figure 2

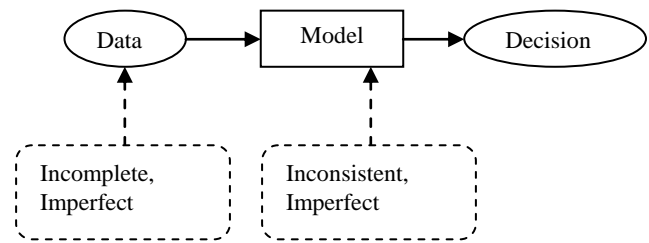

Figure 2 Illustration of the theory of bounded rationality

Figure 2 illustrates that on making decisions rationally certain considerations must be taken into account and these are that the data that is used to make such decisions are never complete and are subject to measurement errors and are thus imperfect and incomplete. The second aspect that should be observed is that the model which takes data and translates these into decisions is also imperfect. If for example this model is a human brain then it is inconsistent in the sense that it may be influenced by other factors such as whether the individual concerned is hungry or angry. Herbert Simon coined a term *satisficing* which is the concept of making an optimized decision under the limitations that the data used in making such decisions are imperfect and incomplete while the model used to make such decisions is inconsistent and imperfect.

Now that we have described the theory of bounded rationality which resulted in the limitations of the applicability of the theory of rational decision making the next section describes the theory of semi-bounded rationality which is theory proposed in this paper.

## 4 Semi-bounded Rational Decision Making

This section proposes the theory of semi-bounded rationality. In order to understand the principle of semi-bounded rationality it is important to state few propositions and these are as follows:
1. Rational decision making is a process of making decisions based on logic and scientific thinking.
2. The process of rationality is indivisible. In other words you cannot be half rational and half irrational. The concept of the indivisibility of rationality and the impact of this is a powerful concept that requires further investigation. Rationality is indivisible because it cannot be broken down into small pieces which when assembled make the whole. There are other concepts that are indivisible such as principle.
3. The theory of bounded rationality does not truncate the theory of rationality but merely specifies the bounds under which the principle of rationality is applied.

With the advances in signal processing, enhanced theories of autoassociative machines and advances in artificial intelligence methods, it has become necessary to revise the theory of bounded rationality. The fact that information which is used to make decisions in imperfect because of factors such as measurement errors can be partially corrected by using advanced data analysis methods and the fact that some of the data that are missing and, thereby, incomplete can be partially completed using missing data estimation methods and the fact that a human brain which is influenced by other social and physiological factors can be substituted for by recently developed artificial intelligence machines implies that the bounds under which rationality is exercised can be shifted and thus bounded rationality can now be viewed as semi-bounded rationality. Tsang (2008) proposed that computational intelligence determines effective rationality. What Tsang implies is that there is a degree of rationality a situation which is only true if rationality is divisible. Rationality cannot be quantified and cannot be sub-divided, it is either there or absent and it only exists as a whole not as a fraction. The model of semi-bounded rationality can thus be expressed as shown in Figure 3.

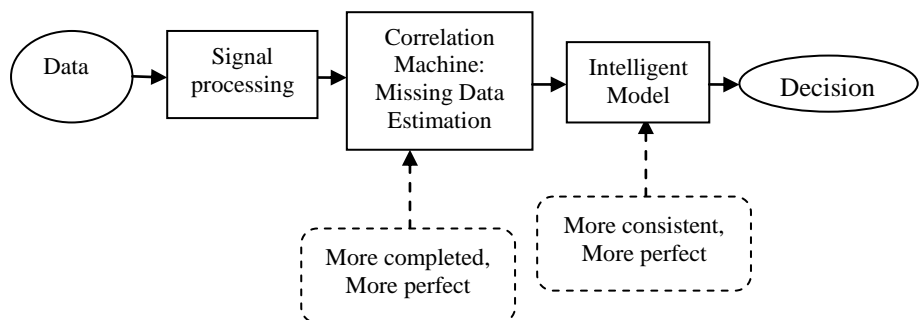

Figure 3 Illustration of the theory of semi-bounded rationality.

The implications of the model in Figure 3 to many disciples such as economics, political science, social science is needs further investigation. It goes to the heart of the relationship between rationality and modern technology. Basically, modern technology now allows us to update the limiting concept of bounded rationality to a less limited concept of semi-bounded rationality. The implication of the concept of the notion of indivisibility of rationality and the application of this profound notion to many areas such as economics, sociology and political science is far reaching and requires further investigation.

## 5 Conclusions

In this paper the concept of semi-bounded rationality is introduced as an extenion to the notion of bounded rationality. It is based on the indivisible principle of rationality but is realized by using advanced signal processing tools, missing data estimation methods and artificial intelligence. Further investigations on the implication of the theory of semi-bounded rationality need further studies.

## References

Adams, R.M. (1976). "Motive Utilitarianism". The Journal of Philosophy 73 (14).

Allingham, M. (2002). Choice Theory: A Very Short Introduction, Oxford.

Anscombe, G. E. M. (January 1958). "Modern Moral Philosophy". Philosophy 33 (124).

Aviad, B., Roy, G. A decision support method, based on bounded rationality concepts, to reveal feature saliency in clustering problems (2012) Decision Support Systems, 54 (1), pp. 292-303.

Chai, J., Liu, J.N.K., Ngai, E.W.T. Application of decision-making techniques in supplier selection: A systematic review of literature (2013) Expert Systems with Applications, 40 (10), pp. 3872-3885.

Badri, N., Rad, A., Kareshki, H., Abdolhamidzadeh, B., Parvizsedghy, R., Rashtchian, D. A risk-based decision making approach to determine fireproofing requirements against jet fires (2013) Journal of Loss Prevention in the Process Industries, 26 (4), pp. 771-781.

Baker, K.M. Decision Making in a Hybrid Organization: A Case Study of a Southwestern Drug Court Treatment Program (2013) Law and Social Inquiry, 38 (1), pp. 27-54.

Bentham, J. (2009). An Introduction to the Principles of Morals and Legislation (Dover Philosophical Classics). Dover Publications Inc.

Bicchieri, C. (2003). "Rationality and Game Theory", in The Handbook of Rationality, The Oxford Reference Library of Philosophy, Oxford University Press.

Foster; K.R. and Kokko H. (2009). "The evolution of superstitious and superstition-like behaviour". Proceedings of the Royal Society B: Biological Sciences 276 (1654): 31–7.

Gama, J. Data stream mining: The bounded rationality (2013) Informatica (Slovenia), 37 (1), pp. 21-25.

Gu, J., Bohns, V.K., Leonardelli, G.J. Regulatory focus and interdependent economic decision-making (2013) Journal of Experimental Social Psychology, 49 (4), pp. 692-698.

Hembacher, E., Ghetti, S. How to bet on a memory: Developmental linkages between subjective recollection and decision making (2013) Journal of Experimental Child Psychology, 115 (3), pp. 436-452.

Jiang, R., Xie, J., Wang, N., Li, J. Evolution game analysis of water saving and pollution prevention for city user groups based on bounded rationality (2013) Shuili Fadian Xuebao/Journal of Hydroelectric Engineering, 32 (1), pp. 31-36.

Jin, C., Jin, S.-W., Tan, H.-Y. Computer virus propagation model based on bounded rationality evolutionary game theory (2013) Security and Communication Networks, 6 (2), pp. 210-218.

Kahneman, D., Tversky, A. (1979)."Prospect Theory: An Analysis of Decision Under Risk". Econometrica, 47 (2), pp. 263–291.

Kahneman, D., (2011). Thinking, Fast and Slow. Macmillan, New York.

Köksalan, M.M. and Sagala, P.N.S., M. M.; Sagala, P. N. S. (1995). "Interactive Approaches for Discrete Alternative Multiple Criteria Decision Making with Monotone Utility Functions". Management Science 41 (7): 1158–1171.

Lee, I.H. Speculation under bounded rationality (2013) Journal of Economic Theory and Econometrics, 24 (1), pp. 37-53.

Mill, J.S. (2011). A System of Logic, Ratiocinative and Inductive (Classic Reprint). Oxford University Press.

Murata, A., Kubo, S., Hata, N. Study on promotion of cooperative behavior in social dilemma situation by introduction of bounded rationality - Effects of group heuristics on cooperative behavior (2012) Proceedings of the SICE Annual Conference, art. no. 6318444, pp. 261-266.

Nozick, R. (1993). The Nature of Rationality. Princeton: Princeton University Press.

Salem, O.M., Miller, R.A., Deshpande, A.S., Arurkar, T.P. Multi-criteria decision-making system for selecting an effective plan for bridge rehabilitation (2013) Structure and Infrastructure Engineering, 9 (8), pp. 806-816.

Serrano González, J., Burgos Payán, M., Riquelme Santos, J. Optimum design of transmissions systems for offshore wind farms including decision making under risk (2013) Renewable Energy, 59, pp. 115-127.

Simon, H. (1957). "A Behavioral Model of Rational Choice", in Models of Man, Social and Rational: Mathematical Essays on Rational Human Behavior in a Social Setting. New York: Wiley.

Simon, H. (1990). "A mechanism for social selection and successful altruism". Science 250 (4988): 1665–8.

Simon, H. (1991). "Bounded Rationality and Organizational Learning". Organization Science 2 (1): 125–134.

Skinner, B. F. (1948). "'Superstition' in the Pigeon". Journal of Experimental Psychology 38 (2): 168–172

Spohn, W. (2002). The Many Facets of the Theory of Rationality. Croatian Journal of Philosophy 2: 247–262.

Stanciu-Viziteu, L.D. The shark game: Equilibrium with bounded rationality (2012) Lecture Notes in Economics and Mathematical Systems, 662, pp. 103-1

Steuer, R.E. (1986). Multiple Criteria Optimization: Theory, Computation and Application. New York: John Wiley.

Tisdell, C. (1996). Bounded Rationality and Economic Evolution: A Contribution to Decision Making, Economics, and Management. Cheltenham, UK: Brookfield.

Tsang, E.P.K. (2008). "Computational intelligence determines effective rationality". International Journal on Automation and Control 5 (1): 63–6

Vyse, S.A. (2000). Believing in Magic: The Psychology of Superstition. Oxford, England: Oxford University Press.

Weber, M. (1922) "Ueber einige Kategorien der verstehenden Soziologie." Pp. 427-74 in Gesammelte Aufsaetze zur Wissenschaftslehre.

Yao, J., Li, D. Bounded rationality as a source of loss aversion and optimism: A study of psychological adaptation under incomplete information (2013) Journal of Economic Dynamics and Control, 37 (1), pp. 18-31.

Zionts, S.; Wallenius, J. (1976). "An Interactive Programming Method for Solving the Multiple Criteria Problem". Management Science 22(6): 652–663.